# Data-Driven Extended Corresponding State Approach for Residual Property Prediction of Hydrofluoroolefins

Gang Wang[a]    Peng Hu[a,*]

[a] Department of Thermal Science and Energy Engineering, University of Science and Technology of China, Hefei 230027, China

**Abstract:**

Hydrofluoroolefins are considered the most promising next-generation refrigerants due to their extremely low global warming potential values, which can effectively mitigate the global warming effect. However, the lack of reliable thermodynamic data hinders the discovery and application of newer and superior hydrofluoroolefin refrigerants. In this work, integrating the strengths of theoretical method and data-driven method, we proposed a neural network extended corresponding state model to predict the residual thermodynamic properties of hydrofluoroolefin refrigerants. The innovation is that the fluids are characterized through their microscopic molecular structures by the inclusion of graph neural network module and the specialized design of model architecture to enhance its generalization ability. The proposed model is trained using the highly accurate data of available known fluids, and evaluated via the leave-one-out cross-validation method. Compared to conventional extended corresponding state models or cubic equation of state, the proposed model shows significantly improved accuracy for density and energy properties in liquid and supercritical regions, with average absolute deviation of 1.49% (liquid) and 2.42% (supercritical) for density, 3.37% and 2.50% for residual entropy, 1.85% and 1.34% for residual enthalpy. These results demonstrate the effectiveness of embedding physics knowledge into the machine learning model. The proposed neural network extended corresponding state model is expected to significantly accelerate the discovery of novel hydrofluoroolefin refrigerants.

---

[*] Corresponding author. E-mail: hupeng@ustc.edu.cn (Peng Hu)

**Nomenclature**

| | | | |
|---|---|---|---|
| *Acronyms* | | $P$ | pressure (MPa) |
| AAD | Average absolute deviation | PFC | perfluorocarbon |
| CFC | chlorofluorocarbons | $r$ | molecular representation vector |
| ECS | extended corresponding state | $R$ | ideal gas constant ($J \cdot mol^{-1} \cdot K^{-1}$) |
| EOS | equation of state | $s$ | dimensionless entropy |
| FCL | fully connected layer | $\mathbf{s}$ | similarity vector |
| GNN | graph neural network | $T$ | temperature (K) |
| GWP | global warming potential | $u$ | dimensionless internal energy \ |
| HC | hydrocarbon | $Z$ | compressibility factor |
| HCFC | hydrochlorofluorocarbon | *Greek symbols* | |
| HFC | Hydrofluorocarbon | $\alpha$ | dimensionless Helmholtz energy |
| HFO | hydrofluoroolefin | $\theta$ | shape factor for temperature |
| LOOCV | leave-one-out cross-validation | $\rho$ | density ($mol \cdot L^{-1}$) |
| N | number of data points | $\phi$ | shape factor for density |
| NN | neural network | $\omega$ | acentric factor |
| PFC | perfluorocarbons | *subscripts* | |
| QSPR | quantitative structure-property relationship | c | critical parameters |
| RB | residual block | cal | calculated value |
| Symbols | | j | target fluid |
| a | molar Helmholtz energy (J · mol$^{-1}$) | o | reference fluid |
| $c_p$ | isobaric heat capacity (J · mol$^{-1}$ · K$^{-1}$) | r | reduced parameters |
| $f_j$ | scaling factor for temperature | ref | reference value |
| g | molar Gibbs energy (J · mol$^{-1}$) | *Superscripts* | |
| h | dimensionless enthalpy | id | ideal gas properties |
| $h_j$ | scaling factor for density | R | residual properties |

## 1. Introduction

Refrigeration and heating cycles are essential to modern life, accounting for a substantial portion of global energy consumption. As the heart of these thermodynamic cycles, refrigerants absorb heat on the lower temperature side and release heat on the higher side to result in the cooling or heating effect. In the past decades, refrigerants have gone through a continuous evolution in response to increasingly serious environmental problems [1]. Some early refrigerants, such as chlorofluorocarbons (CFC) or hydrochlorofluorocarbons (HCFC), have been phased out due to their damage to the ozone sphere. Hydrofluorocarbons (HFC) were once widely used as the replacement for CFC and HCFC, but are now being phased out due to their high global warming potential (GWP) values. Currently, hydrofluoroolefins (HFO) are regarded as the most promising alternatives to these old refrigerants because of their extremely low GWP values and no harm to the ozone sphere.

Accurate thermodynamic data of HFO refrigerants is the foundation of cycle simulation and optimization. Since the 2010s, researchers have conducted numerous experimental measurements on the thermodynamic properties of HFO refrigerants [2], including their critical point [3; 4], vapor pressure [5; 6], $PVT$ [7; 8], speed of sound [9], and so on. However, obtaining the accurate and complete thermodynamic properties is a time-consuming and labor-intensive task: a large number of data points should be measured for each different property, then these data should be comprehensively evaluated to fit the multiparameter equation of state (EOS), and this process usually takes several years. Up to now, only eight HFO refrigerants [10-17] have accurate Helmholtz energy equation of state, there are still a large number of HFO compounds whose thermodynamic properties are unknown.

The lack of reliable thermodynamic data has been an obstacle to the discovery of newer and superior HFO refrigerants. In this case, some semi-empirical methods, such as the PR [18] equation of state or the extended corresponding state (ECS) model with the empirical formulas [19; 20], can provide preliminary predictions. These semi-empirical models require only a fluid's critical parameters and acentric factor ($\omega$) to perform calculations, eliminating the need for extensive experimental data fitting. The incorporation of certain theoretical considerations gives them some degree of predictive ability, but the empirical mathematical form also brings some limitations. For instance, the cubic EOS is known to have a large deviation in the liquid phase. For the ECS model with empirical formulations, it is demonstrated to be accurate if a fluid-specific ECS model that fits only a few saturation [21; 22] or single-phase [23-25] data is used; nevertheless, the universal ECS model usually shows relatively large deviations for fluids with only the critical parameters and acentric factor available.

In recent years, data-driven approach has been widely used to calculate the thermophysical properties of fluids [26-28]. Noushabadi et al. [29] used four different machine learning methods to estimate saturated vapor enthalpy, entropy, sound speed, and viscosity of refrigerants, where the input variables of model are critical parameters, temperature, and pressure. Rathod et al. [30] also used a similar approach to predict the thermophysical properties of fluids but the critical parameters in input variables are replaced by molecular descriptors. However, pure data-driven approach was employed in their work and no theoretical knowledge was incorporated into the model. In engineering applications with limited reliable data, the data-driven approach must be combined with certain physical knowledge to make sure it learns the true principle rather than over-fitting to the dataset. Several methods already exist to integrate physical knowledge with machine models [31], such as using machine learning model to predict the parameters of existing thermodynamic models rather than replace the full thermodynamic models [32-35], using the data generated from physics model to pre-train the machine learning model and then fine-tune it using accurate experimental data [36; 37].

In this study, extended corresponding stated theory [38] was integrated with neural network to establish an accurate thermodynamic prediction model for HFO compounds, aiming to accelerate the discovery of new HFO refrigerants and push their applications. With the main target being HFO compounds, HFC and hydrocarbon (HC) compounds are also included in this work to expand the dataset. Indeed, there have been studies [39; 40] that employed neural network to formulate the shape factors in ECS model; however, limited by the historical reasons and the amount of available data, their work only developed the ECS model for some already investigated fluids and did not develop a universal ECS model for unknown fluids. Fortunately, with the rapid development of advanced deep learning techniques and the availability of thermodynamic data for numerous compounds, developing a more accurate universal ECS model has now become feasible. This work is the first time that use the novel deep learning technique to develop a universal ECS model for HFO compounds, and we refer to the proposed model as neural network extended corresponding state (NN-ECS) model. In this study, leveraging the powerful fitting capability of neural network, highly accurate data for a large number of fluids in the REFPROP database [41] was used in the model development. Moreover, the NN-ECS model further takes into account the molecular structure similarity between the reference fluid and the target fluids by the reasonable inclusion of a graph neural network (GNN) module. Leave-one-out cross-validation (LOOCV) was used to evaluate the predictive ability of model and to avoid the over-optimistic results. Compared with the conventional ECS models or the cubic EOS, the proposed NN-ECS significantly improves the prediction accuracy for density and energy properties in the liquid and supercritical phases. The proposed NN-ECS model is a reliable tool for the estimation of thermodynamic properties of HFO compounds, and we believe that it will play an important role in the discovery of new environmentally friendly HFO refrigerants and in further reducing the global warming effect.

## 2. Methods

*2.1 Extended corresponding state theory*

Currently, the EOS of fluids is typically expressed in its Helmholtz energy. The advantage of using Helmholtz energy is that all other thermodynamic properties can be derived by differentiating the Helmholtz energy, thereby eliminating the complex integration operation and enabling the use of multi-type experimental values in EOS

development [42]. The Helmholtz energy is divided into two parts:

$$\alpha = \frac{a}{RT} = \alpha^{id} + \alpha^{R} \tag{1}$$

where $a$ is the molar Helmholtz energy, $R$ is the ideal gas constant, $\alpha$ is the dimensionless Helmholtz energy, $\alpha^{id}$ is the ideal gas contribution that could be obtained from the ideal gas isobaric heat capacity ($c_p^{id}$), and $\alpha^R$ is the residual part that comes from the intermolecular interactions. $\alpha^R$ is usually determined by fitting to experimental data of multi-type properties.

The corresponding state theory [38] states that two fluids have the same dimensionless residual Helmholtz energy if they have the same reduced intermolecular potential, as in Eq. (2):

$$\alpha_j^R(T_j, \rho_j) = \alpha_o^R(T_o, \rho_o) \tag{2}$$

where subscript $j$ and $o$ indicate the target and reference fluids, respectively. The temperature and density of target fluid ($T_j, \rho_j$) and reference fluid ($T_o, \rho_o$) have the following relationship:

$$T_o = T_j/f_j, \rho_o = \rho_j h_j \tag{3}$$

where $f_j$ and $h_j$ are scaling factors, they are related to the ratio of critical parameters:

$$f_j = T_{c,j}/T_{c,o}, h_j = \rho_{c,o}/\rho_{c,j} \tag{4}$$

However, the corresponding state theory is only valid for simple spherically symmetric molecules. Shape factors are introduced to make the method applicable to non-spherical molecules, this is known to be the extended corresponding state method:

$$f_j = T_{c,j}/T_{c,o}\,\theta_j(T_j, \rho_j) \tag{5a}$$

$$h_j = \rho_{c,o}/\rho_{c,j}\, \phi_j(T_j,\rho_j) \tag{5b}$$

where $\theta_j$ and $\phi_j$ are the shape factors, which are dependent on $T_j$ and $\rho_j$.

Even though the ECS theory has a good foundation in molecular theory, it is difficult to determine the shape factors directly from theory method; indeed, shape factors are usually obtained by fitting to macroscopical experimental data. Once the shape factors are known, the residual Helmholtz energy of target fluids can be represented by that of the reference fluid. Subsequently, other thermodynamic properties of the target fluids, e.g., *PVT*, vapor pressure, and residual enthalpy, can be derived by differentiating the residual Helmholtz energy. The thermodynamic properties of target and reference fluids have the following relationship:

$$Z_j^R = u_o^R F_\rho + Z_o^R(1 + H_\rho) \tag{6a}$$

$$u_j^R = u_o^R(1 - F_T) - Z_o^R H_T \tag{6b}$$

$$s_j^R = s_o^R - u_o^R F_T - Z_o^R H_T \tag{6c}$$

$$h_j^R = h_o^R + u_o^R(F_\rho - F_T) + Z_o^R(H_\rho - H_T) \tag{6d}$$

where $Z^R$ is the residual compressibility factor, $u^R$, $s^R$, and $h^R$ are the dimensionless residual internal energy, dimensionless residual entropy, and dimensionless residual enthalpy, respectively; $F_T$, $F_\rho$, $H_T$, and $H_\rho$ are calculated from the scaling factors:

$$F_T = \left(\frac{\partial f_j}{\partial T_j}\right)_{\rho_j}\left(\frac{T_j}{f_j}\right), F_\rho = \left(\frac{\partial f_j}{\partial \rho_j}\right)_{T_j}\left(\frac{\rho_j}{f_j}\right) \tag{7a}$$

$$H_T = \left(\frac{\partial h_j}{\partial T_j}\right)_{\rho_j}\left(\frac{T_j}{h_j}\right), H_\rho = \left(\frac{\partial h_j}{\partial \rho_j}\right)_{T_j}\left(\frac{\rho_j}{h_j}\right) \tag{7b}$$

*2.2 Neural network extended corresponding state (NN-ECS) model*

The most important part of the ECS method is how to formulate the shape factors of target fluids relative to the reference fluid, and the previous studies [19; 20] all used empirical methods. In this work, data-driven approach was employed to discover the hidden relationship for shape factors by automatically learning from a larger amount of data, aiming to avoid the possible limitations introduced by the manually designed formula. The architecture of NN-ECS model is shown in Fig. 1. It is made up of two main components: one is a GNN module used to calculate the molecular structure similarity, and the other one is a multilayer feedforward neural network for shape factors.

Graph neural network is capable of generating high-quality representations of molecules by learning from their molecular structure. GNN treats the molecule as an undirected graph and then generates its representation vector by conducting the aggregation, update, and readout operations on it. The atoms and bonds in the molecule correspond to the nodes and edges in the graph, respectively, and they are characterized by feature vectors in Table 1 and Table 2. In contrast to the manually-designed molecular descriptors, GNN contains adjustable parameters thus the learned molecular representation depends on the special task instead of fixed values. The GNN model developed by Xiong et al. [43] was employed in this work, and it has been included in the DGl-LifeSci package [44]. The molecular graph of the reference fluid and target fluids are input into the GNN module to generate their molecular representations, i.e.,

$r_o$ and $r_j$, and then the similarity vector is calculated as:

$$s = \frac{r_o \odot r_j}{\|r_o\|\|r_j\|} \tag{8}$$

Eq. (8) is similar to the formula of cosine similarity, with the difference being that the Hadamard product is used in the numerator rather than the dot product. The reason is that the Hadamard product is in an element-wise manner so that vector $s$ contains richer similarity information in each dimension, whereas the dot product only obtains a scalar value for similarity.

Table 1 Feature vector used to represent the atom state.

| Atom feature | Size | Description |
| --- | --- | --- |
| Atom type | 2 | (C F) [one-hot] |
| Degree | 4 | Number of covalent bonds (1,2,3,4) [one-hot] |
| Hybridization | 2 | ($sp^2$, $sp^3$) [one-hot] |
| Hydrogens | 4 | Number of connected hydrogens (0,1,2,3) [one-hot] |

Table 2 Feature vector used to represent the bond state.

| Bond feature | Size | Description |
| --- | --- | --- |
| Bond type | 2 | (single, double) [one-hot] |
| Stereo | 3 | (StereoNone, StereoZ, StereoE) [one-hot] |

The second component is a multilayer feedforward neural network for shape factors, with the inputs being reduced temperature ($T_r = T/T_c$) and reduced density ($\rho_r = \rho/\rho_c$). The input layer is fed into a fully connected layer (FCL1) to reshape the input size from 2 to 32, and then it is followed by two residual blocks (RB1 and RB2) [45] to increase the depth of the network. Half of the output of the second residual block undergoes Hadamard product with vector $\mathbf{s}$ to incorporate the information of molecular structure similarity. Finally, the shape factors are obtained by passing through a fully connected layer (FCL2).

Note that the target fluids studied in this work are limited to hydrocarbon or fluorinated hydrocarbon, thus the trends of shape factors are very similar and do not change drastically with the variation of fluids; indeed, it is the state variables ($T_r$ and $\rho_r$) that have the greatest influence on the shape factors. In order to avoid over-fitting, the emphasis of model should be put on learning from $T_r$ and $\rho_r$, while the similarity vector $\mathbf{s}$ should play a less significant role in the model. Two approaches are adopted to achieve this:

- Vector $\mathbf{s}$ is not concatenated with ($T_r$, $\rho_r$) as joint input; indeed, only $T_r$ and $\rho_r$ are used as initial inputs, thus the FCL1, RB1, and RB2 layers are expected to learn the universal trends for shape factors. The similarity vector $\mathbf{s}$ is incorporated at later layers so that only layer FCL2 without the activation function is responsible for capturing the difference among target fluids.

- The dimension of $\mathbf{s}$ is only half that of RB2's output; consequently; only half the parameters in the FCL2 layer are relevant to the type of fluids while the other half remain universal and are solely related to $T_r$ and $\rho_r$. This design further reduces the dependence of the calculated shape factors on the type of fluids.

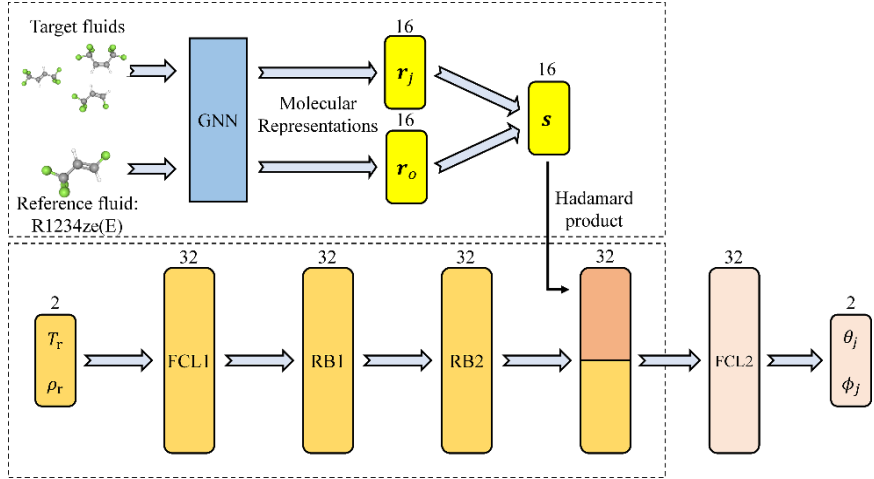

Fig. 1. Architecture of the neural network extended corresponding state (NN-ECS) model.

*2.3 Model training and evaluation*

As mentioned above, once the shape factors are obtained, the thermodynamic properties of the target fluid can be fully represented by those of the reference fluid. The model is trained by minimizing the following loss function:

$$loss = \frac{1}{M \times N}\sum_{m=1}^{M}\sum_{n=1}^{N}(|\alpha_{m,n,cal}^{R} - \alpha_{m,n,ref}^{R}| + |Z_{m,n,cal}^{R} - Z_{m,n,ref}^{R}| + |u_{m,n,cal}^{R} - u_{m,n,ref}^{R}|) \quad (9)$$

there are two summations where the first is for fluids and the second is for state points; $\alpha^R$, $Z^R$, and $u^R$ are dimensionless residual Helmholtz energy, residual compressibility factor, and dimensionless residual internal energy, respectively; the subscript $cal$ and $ref$ indicate the value calculated by NN-ECS and the reference value in the REFPROP database [41]. As specified in Eq. (6) and Eq. (7), calculating the residual thermodynamic properties of target fluids requires the derivative of shape factors $\theta_j$ and $\phi_j$ with respect to $T_j$ and $\rho_j$. These derivates are computed using the automatic differentiation technique in PyTorch.

Considering that fluids exhibit similar but not identical behavior at reduced temperature and density, the shape factors are values close to but not equal to 1. Furthermore, note that this is an under-determined problem as there are two variables ($\theta_j$ and $\phi_j$) but only one equation (i.e., Eq. (1)) exists. The output of a randomly initialized model typically deviates significantly from 1, causing training failure if directly minimizing the loss function in Eq. (9). More specifically, the randomly initialized parameters have a huge distance from the true solution, thus the gradient-based algorithms will find wrong optimization directions and lead to worse solution step by step, i.e., the model output will be further and further away from 1 until it becomes negative or very large values, that is obviously wrong in physics. A solution to this issue is to introduce an arbitrary restriction, such as $u_o^R F_\rho + Z_o^R H_\rho = 0$ [46], thus this becomes a determined problem with two variables and two equations. Here another solution was adopted in this study, we first pretrain the model using the following loss function:

$$loss\_pre = \frac{1}{M \times N} \sum_{m=1}^{M} \sum_{n=1}^{N} (|\theta_j - 1| + |\phi_j - 1|) \qquad (10)$$

This pretraining phase makes the model equivalent to the simple corresponding state model in Eq. (4) and makes the model output equal to 1. Thus, bringing the parameters of NN-ECS very close to the real solution. In this case, the gradient-based optimization algorithms can find the right direction to minimize the loss function in Eq. (9), and the convergency is guaranteed.

A total of 44 fluids were used in the model training, comprising 20 HC, 14 HFC, 6 HFO, and 4 perfluorocarbons (PFC). These fluids can be considered to have been well-studied, and accurate EOS are available for them in the REFPROP database. Almost all EOS are explicit in Helmholtz energy except a few are explicit in pressure; the specific EOS for each fluid can be found in this publication [42]. Furthermore, for the three HFO refrigerants (R1123 [10], R1234yf [15], and R1243zf [14]), we employed their newly developed Helmholtz EOS to calculate their properties. The EOS in REFPROP, including the three new ones, could be considered the most accurate thermodynamic property resources, as they are developed by assessing and fitting extensive multi-type experimental data within their uncertainties and could extrapolate to the region with no data through proper restrictions. Taking HFO refrigerants as an example: the vapor pressure is usually calculated with an uncertainty of 0.1%; the liquid density is usually calculated with an uncertainty of 0.1%, the uncertainty in vapor density is usually higher with a typical value of less than 0.5%, and the uncertainty in density always increases as close to the critical point; energy properties, such as the entropy and enthalpy, are usually derived from the correlation of both sound speed and *PVT* data, the sound speed data is usually calculated with an uncertainty of 0.05% or less. As most experimental exploration is focused on the liquid and vapor phase, the experimental data in the supercritical phase are few in the development of EOS; therefore, considering the good extrapolation performance of Helmholtz EOS, its uncertainties in supercritical phase should be larger than those in the liquid phase.

The training data were generated using the REFPROP software. Considering the different applicable ranges of each fluid's EOS, a unified range was applied to generate training data, as shown in Table 3, to ensure that this range covers the applicable range of most fluids. In this unified range, the minimum and maximum temperatures are $0.7T_c$ and $1.1T_c$, with a step of 10 K. For pressure, the minimum is 0.1 MPa, the maximum is 50 MPa for the liquid phase, and a narrow pressure range with a maximum of $5P_c$ was adopted for the supercritical region considering that fewer experimental data in this region was available in the development of most EOS. A pressure step of 0.2 MPa is applied for $P < P_c$, considering that most refrigeration or heating cycles operate in this region; a step of 3 MPa is applied for the condition of $P > P_c$ and $P_{max} < 50$ MPa in the liquid phase; a step of 6 MPa is applied for the condition of $P > P_c$ and $P_{max} = 50$ MPa in liquid phase; and a step of 2 MPa is applied for the supercritical phase. Approximately 430 state points were generated for each fluid.

Table 3 Temperature and pressure ranges with steps for state point generation.

| | Minimum | Step | Maximum[a] |
|---|---|---|---|
| Temperature (K) | $0.7T_c$ | 10 K | $1.1T_c$ |
| Pressure (MPa) | 0.1 MPa | 0.2 MPa for $P < P_c$, <br> 3 MPa for $P > P_c$ and $P_{max} < 50$ MPa <br> 6 MPa for $P > P_c$ and $P_{max} = 50$ MPa <br> 2 MPa in supercritical phase | 50 MPa in liquid phase <br> $5P_c$ in supercritical phase |

[a] If the maximum pressure of EOS is less than 50 MPa or $5P_c$, use it instead of 50 MPa or $5P_c$

The goal is to obtain a model that is able to accurately predict the properties of unknown fluid, rather than only fitting the existing data very well. Hence evaluating the model's generalization ability to new fluid becomes a vital part. It is not practical to split the dataset into training and test sets at a ratio of 7:3, owing that only 44 fluids are included in the dataset and this splitting means that 30 fluids are used in model training, this is a waste of available data. In this case, leave-one-out cross-validation (LOOCV) was employed to evaluate the model performance, as shown in Fig. 2. In each iteration, one fluid is selected as the test set while the remaining 43 fluids are training set; the NN-ECS is trained in the training set and then report its result in the test set. Repeat this process until each fluid has been used as the test set. In the following sections, when presenting the prediction result for a given fluid, the LOOCV method ensures the exclusion of its data from model training, thereby providing a more rigorous evaluation of the model's predictive performance for unknown fluids. We refer to the models trained using the LOOCV methods as the LOOCV models, they are only used for model evaluation. In addition, we provided a final model that is trained using the full-data of 44 fluids, and this final model will be used in real predictions.

The average absolute deviation (AAD) is given by:

$$AAD = \frac{100}{N} \sum_{i=1}^{N} \left| \frac{X_{i,ref} - X_{i,cal}}{X_{i,ref}} \right| \quad (11)$$

where $X_{i,ref}$ is the reference value in REFPROP, and $X_{i,cal}$ is the value estimated by models.

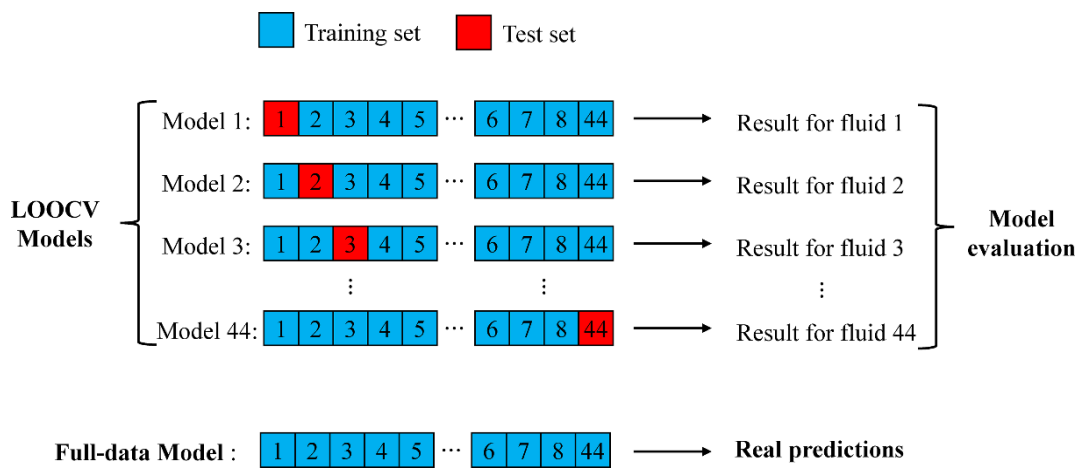

Fig. 2. Illustration of the leave-one-out cross-validation (LOOCV) method for model training and evaluation.

The model is trained using an NVIDIA RTX 4060 Ti GPU, with one iteration taking approximately one hour. Once the training is finished on the dataset, the NN-ECS can predict the properties of unknown HFO or HFC fluids very quickly through forward propagation, whether running on CPU or GPU. However, since the NN-ECS relies on automatic differentiation technique instead of explicit formulas to calculate derivatives, its computational time is longer than that of multiparameter equation of state.

**3.Results**

*3.1 Selection of the reference fluid*

In this work, four fluids (propane [47], R143a [48], R1234yf [15], and R1234ze(E) [17]) were selected to explore the influence of different reference fluids. Each of them was used as the reference fluid in turn and then LOOCV was performed to test the models' generalization ability. The AAD in single phase properties prediction, i.e., density, residual entropy ($s^R$), and residual enthalpy ($h^R$), are presented in Table 4. For density predictions, the AAD exhibits minimal variations (< 0.7%) across the four reference fluids in both phases. Similarly, small variations were observed for $s^R$ and $h^R$ in liquid and supercritical phases, demonstrating the model's robustness to the changes of reference fluid. Although the studied fluids are HFC and HFO, using propane as the reference fluid achieves the same accuracy as using HFC or HFO. This could be attributed to the strong fitting capability of neural network, which can discover the difference between the intermolecular potential of propane and the studied fluids by learning from the available dataset, and then making accurate predictions for the HFO and HFC not in the dataset. Considering the effect of randomness in the training process, it is concluded that the reference fluid choice has a negligible impact on model's generalization ability. In the following sections, R1234ze(E) was employed as the reference fluid for all results.

Table 4 AAD (%) in single phase properties prediction across four different reference fluids

|  | Liquid | | | Gas | | | Supercritical | | |
| --- | --- | --- | --- | --- | --- | --- | --- | --- | --- |
|  | $\rho$ | $s^R$ | $h^R$ | $\rho$ | $s^R$ | $h^R$ | $\rho$ | $s^R$ | $h^R$ |
| Propane | 1.57 | 3.07 | 1.73 | 0.84 | 8.49 | 5.43 | 2.65 | 2.48 | 1.65 |
| R143a | 1.23 | 2.89 | 1.64 | 0.80 | 7.98 | 5.30 | 2.00 | 2.74 | 1.33 |
| R1234yf | 1.60 | 2.73 | 2.73 | 1.05 | 10.10 | 6.63 | 2.62 | 2.96 | 2.02 |
| R1234ze(E) | 1.49 | 3.37 | 1.85 | 0.73 | 7.49 | 4.69 | 2.42 | 2.50 | 1.34 |

*3.2 Prediction result on multiply thermodynamic properties*

Using the values in REFPROP as the reference values, the prediction performance of the developed NN-ECS model is evaluated in various thermodynamic properties, including vapor pressure, *PVT*, residual enthalpy, and residual entropy.

The NN-ECS model is compared with several existing models. The first is the ECS model developed by Huber and Ely [19], where the shape factors are formulates as follows:

$$\theta_j = 1 + (\omega_j - \omega_o)(\alpha_1 + \alpha_2 \ln T_{r,j}) \tag{12a}$$

$$\phi_j = \frac{Z_{c,o}}{Z_{c,j}}[1 + (\omega_j - \omega_o)(\beta_1 + \beta_2 \ln T_{r,j})] \tag{12b}$$

The formulations contain four parameters ($\alpha_1$, $\alpha_2$, $\beta_1$, and $\beta_2$), and $Z_c$ is the critical compressibility factor. For a given fluid, this method achieves high accuracy in calculating both the saturation and single-phase properties when using fluid-specific parameters, that are fitted from its vapor pressure and saturated liquid density data. In order to make this model applicable to unknown fluids, Huber and Ely proposed the universal parameters by correlating nine refrigerants with R134a as the reference fluid, as shown in Table 5.

The second is the ECS model developed by Teraishi et al. [20], their model formula is consistent with the previous one except that the parameters are fitted from eight new HFO and hydrochlorofluoroolefins (HCFO) refrigerants, they provided two version of models, we choose the one that adopted R1234ze(E) as reference fluid, and the universal parameters are also given in Table 5.

Furthermore, the method of Huber and Ely was refitted to the dataset in this work to enable a fair comparison. Both the fluid-specific parameters (for each studied fluid) and the universal parameters were obtained; the former are given in supplementary materials, while the latter are presented in Table 5.

The last baseline model is the PR EOS [18], which is one of the most widely used cubic EOS. Note that while the dataset comprises HFO, HFC, PFC, and HC, only results for HFO, HFC, and PFC are reported in the subsequent sections, considering that this study is focused on fluorine-containing refrigerants and HC is only used to expand the scope of dataset.

For all the models mentioned above, the full-data version of NN-ECS model and the ECS model using fluid-specific parameters were categorized as Group 1, since single-phase or saturation data are directly incorporated into the model development. Consequently, their results cannot reflect generalization capability for unknown fluids. The remaining models, including NN-ECS trained via LOOCV, the ECS model with universal parameters, and the PR EOS, were classified as Group 2. These models rely solely on critical parameters and acentric factors for calculations, thereby demonstrating true generalization performance for unknown fluids.

Table 5 Universal parameters for the ECS formulation proposed by Huber and Ely

| Reference fluid | R134a (Huber and Ely) | R1234ze(E) (Teraishi et al.) | R1234ze(E) (This work) |
|---|---|---|---|
| $\alpha_1$ | 0.086853583565 | 0.06234 | 0.03458966 |
| $\alpha_2$ | -0.55945094628 | -0.6471 | -0.65863522 |
| $\beta_1$ | 0.057382113745 | -0.3790 | -0.20994404 |
| $\beta_1$ | 0.20164093938 | 0.1040 | 0.16257211 |
| Comment | Fitted to 9 HCFC and HFC | Fitted to 8 HCFO and HFO | Fitted to the dataset in this work |

*3.2.1 Vapor pressure*

In the vapor-liquid equilibrium, the temperature, pressure, and Gibbs free energy of vapor and liquid phases are equal:

$$T_l = T_v \tag{13a}$$

$$P_l = P_v \tag{13b}$$

$$g_l = g_v \tag{13c}$$

the equality of Gibbs free energy can be expressed as follows:

$$g_l^r + \ln \rho_l = g_v^r + \ln \rho_v \tag{13d}$$

where $g_l^r$ and $g_v^r$ are residual Gibbs energy of saturated liquid and saturated vapor, $\rho_l$ and $\rho_v$ are the density of saturated liquid and saturated vapor, respectively. The value of vapor pressure can be obtained by finding the solution that satisfies the above equality conditions [49].

The detailed AAD of vapor pressure for each fluid is provided in supplementary materials. Here only the statistics results in the total dataset are provided in Table 6. For the full-data NN-ECS, it can accurately represent the vapor pressure of fluids with an

AAD of 0.64% although the vapor pressure data was not directly used in model fitting. This is because the full-data model accurately represents the Helmholtz energy surface of each fluid by learning from their single-phase data. However, the prediction performance of NN-ECS is relatively poor with an AAD of 3.76% in the total dataset and a maximum AAD of 9.30% for R161, as seen in the results of LOOCV models. The ECS model developed by Teraishi et al. and PR EOS yield the most accurate vapor pressure predictions, with AAD of 0.24% and 0.35% in the total dataset, respectively; and they maintain AAD below 0.5% for most fluids, demonstrating excellent prediction accuracy for vapor pressure.

Table 6 AAD (%) of vapor pressure prediction results among different models.

|  | NN-ECS (full-data model) | NN-ECS (LOOCV models) | ECS (Teraishi et al.) | ECS (Huber and Ely) | PR-EOS |
|---|---|---|---|---|---|
| Max | 1.84 | 9.30 | 0.87 | 4.90 | 0.71 |
| Mean | 0.64 | 3.76 | 0.20 | 2.02 | 0.33 |

*3.2.2 Single-phase density*

The results for the single-phase density of different methods are presented in Table 7. Furthermore, Fig. 3 presents the AAD values for each individual HFO and HFC fluids across three models: the NN-ECS, the ECS with universal parameters determined by Huber and Ely, and the PR EOS. From the result of LOOCV models, it is observed that the proposed NN-ECS significantly improves the prediction accuracy for the density of liquid and supercritical phases, with AADs of 1.49% and 2.42%, respectively, whereas other models exhibit substantially higher AADs exceeding 4% and 3.2%, respectively. Moreover, the NN-ECS model achieves a significantly lower maximum AAD compared to other models, demonstrating its reliable capabilities in liquid density predictions. For the density of gas phase, the AADs show minor variations across different models, with all of them having an AAD of less than 1%. The reason is that the behavior in the gas phase is simple due to the weak intermolecular interactions thus it is easy to predict.

Table 7 AAD (%) of single-phase density prediction results across different models.

|  | Liquid | | Vapor | | Supercritical | |
| --- | --- | --- | --- | --- | --- | --- |
|  | Mean | Max | Mean | Max | Mean | Max |
| Group 1 | | | | | | |
| NN-ECS (full-data model) | 0.17 | 0.65 | 0.29 | 0.84 | 0.70 | 1.84 |
| ECS (fluid-specific parameters, this work) | 0.63 | 1.63 | 0.79 | 2.02 | 1.29 | 2.70 |
| Group 2 | | | | | | |
| NN-ECS (LOOCV models) | 1.49 | 4.53 | 0.73 | 2.13 | 2.42 | 6.48 |
| ECS (Huber and Ely) | 4.06 | 9.22 | 0.81 | 1.33 | 3.96 | 6.63 |
| ECS (Teraishi et al.) | 4.79 | 9.65 | 0.57 | 1.50 | 3.60 | 7.82 |
| ECS (universal parameters, this work) | 4.33 | 9.62 | 0.58 | 1.47 | 3.21 | 7.34 |
| PR-EOS | 4.14 | 8.99 | 0.95 | 1.51 | 5.62 | 9.82 |

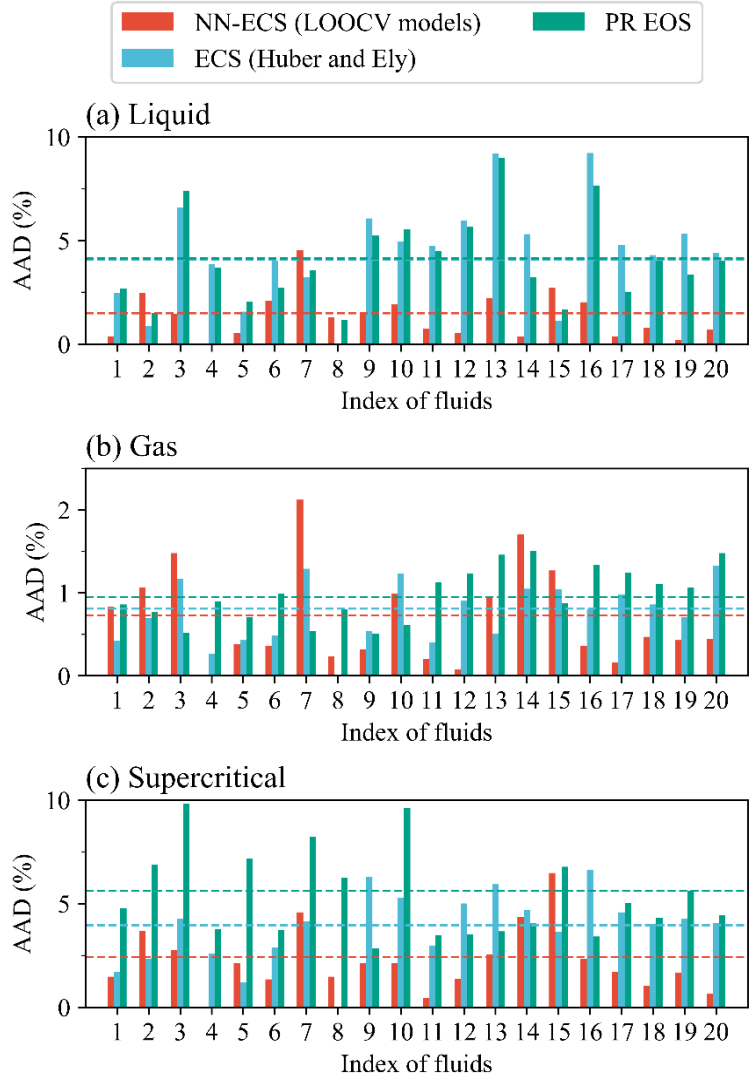

Fig. 3. AAD of the single-phase density for each fluid across different phases: (a) Liquid, (b) gas, (c) Supercritical. The dashed lines indicate the AAD in total dataset.

*3.2.3 Residual entropy*

The results for the residual entropy are presented in Table 8 and the AAD for each individual fluid is shown in Fig. 4. In the liquid and supercritical regions, the proposed

NN-ECS achieves the lowest AAD values of 3.37% and 2.50%, respectively, demonstrating superior prediction performance compared to other models that exhibit AAD exceeding 6% and 4.5%, respectively. Furthermore, the proposed NN-ECS demonstrates significantly lower maximum AAD values compared to other models, with values of 8.4% for liquid and 5.22% for supercritical phase, whereas other models exhibit maximum AAD exceeding 14% and 8%. In the gas phase, most models show significant deviations for residual entropy. Notably, the NN-ECS model even fails to correlate these data accurately, as evidenced by the result of the full-data model. This phenomenon results from the small proportion of residual entropy in the gas phase, and will be discussed in detail in section 4.1.

Table 8 AAD (%) of residual entropy prediction results among different models.

|  | Liquid | | Vapor | | Supercritical | |
| --- | --- | --- | --- | --- | --- | --- |
|  | Mean | Max | Mean | Max | Mean | Max |
| Group 1 | | | | | | |
| NN-ECS (full-data model) | 0.55 | 1.91 | 5.34 | 18.18 | 1.31 | 3.74 |
| ECS (fluid-specific parameters, this work) | 2.65 | 6.87 | 9.23 | 19.83 | 2.87 | 5.65 |
| Group 2 | | | | | | |
| NN-ECS (LOOCV models) | 3.37 | 8.40 | 7.49 | 18.87 | 2.50 | 5.22 |
| ECS (Huber and Ely) | 8.08 | 15.52 | 7.24 | 24.8 | 4.74 | 8.3 |
| ECS (Teraishi et al.) | 10.53 | 24.93 | 6.36 | 14.77 | 6.04 | 12.78 |
| ECS (universal parameters, this work) | 9.75 | 24.49 | 6.33 | 15.34 | 5.55 | 12.25 |
| PR-EOS | 6.29 | 14.57 | 18.05 | 26.08 | 5.53 | 13.92 |

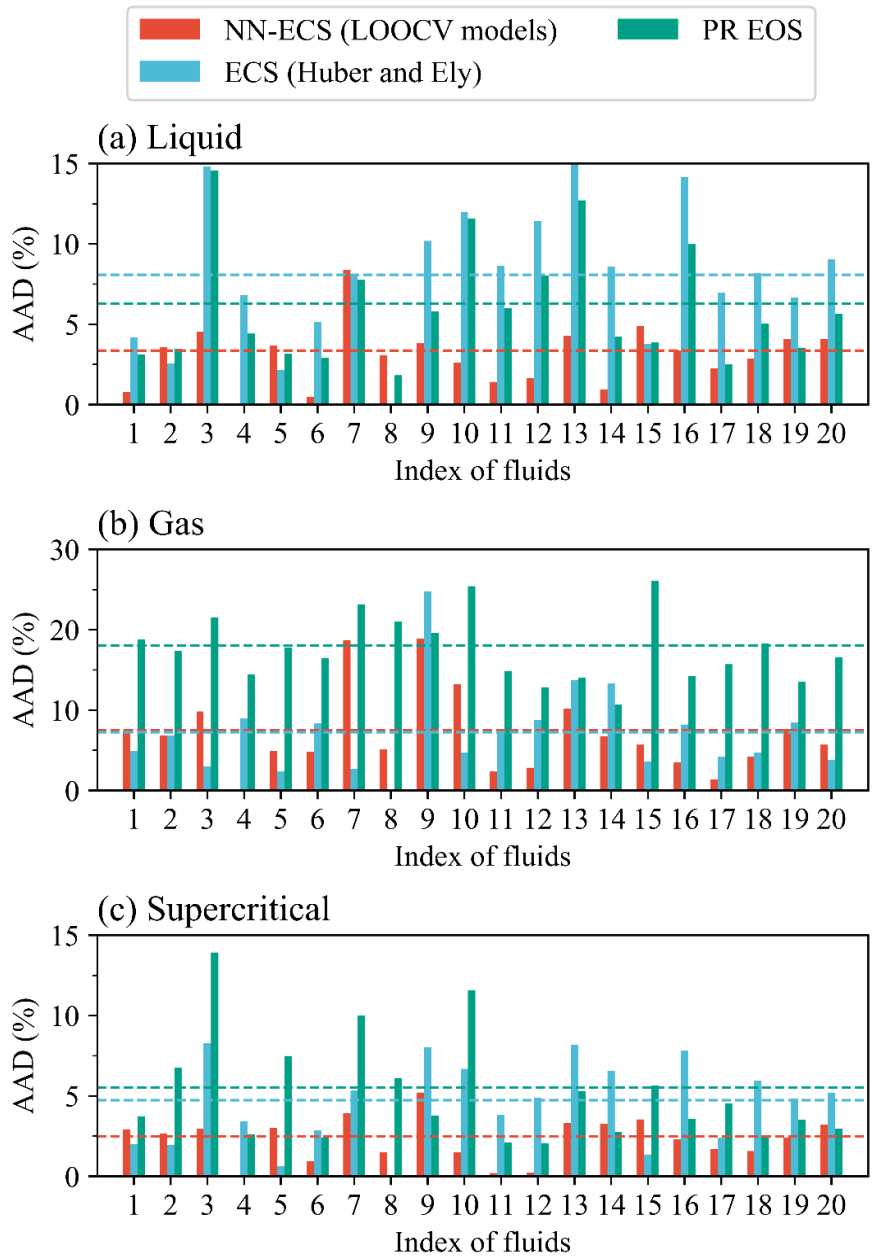

Fig. 4. AAD of the residual entropy for each fluid across different phases: (a) Liquid,

(b) gas, (c) Supercritical. The dashed lines indicate the AAD in total dataset.

*3.2.4 Residual enthalpy*

The results for the residual enthalpy are given in Table 9, and the AAD for each fluid is shown in Fig. 5. In the liquid phase, the NN-ECS model has the lowest AAD of 1.85%, and that of PR EOS is also relatively low, with AAD of 2.73%. In the supercritical region, all the prediction models show an AAD of less than 1.5%, and the PR EOS gives the best results with an AAD of 1.11%. Similar to the result of residual entropy, most models show large AAD for the residual enthalpy in the gas phase, and this has almost no influence on the calculation of total enthalpy.

Table 9 AAD (%) of residual enthalpy prediction results among different models.

|  | Liquid | | Vapor | | Supercritical | |
| --- | --- | --- | --- | --- | --- | --- |
|  | Mean | Max | Mean | Max | Mean | Max |
| Group 1 | | | | | | |
| NN-ECS (full-data model) | 0.22 | 0.60 | 2.55 | 6.05 | 0.58 | 1.51 |
| ECS (fluid-specific parameters, this work) | 0.43 | 0.9 | 5.50 | 13.65 | 1.05 | 1.89 |
| Group 2 | | | | | | |
| NN-ECS (LOOCV models) | 1.85 | 8.40 | 4.69 | 14.11 | 1.34 | 2.85 |
| ECS (Huber and Ely) | 3.25 | 7.19 | 4.63 | 10.11 | 1.24 | 2.45 |
| ECS (Teraishi et al.) | 5.64 | 16.39 | 3.58 | 8.13 | 1.28 | 2.62 |
| ECS (universal parameters, this work) | 5.14 | 14.83 | 3.55 | 8.01 | 1.19 | 2.37 |
| PR-EOS | 2.73 | 7.00 | 7.62 | 11.71 | 1.11 | 2.88 |

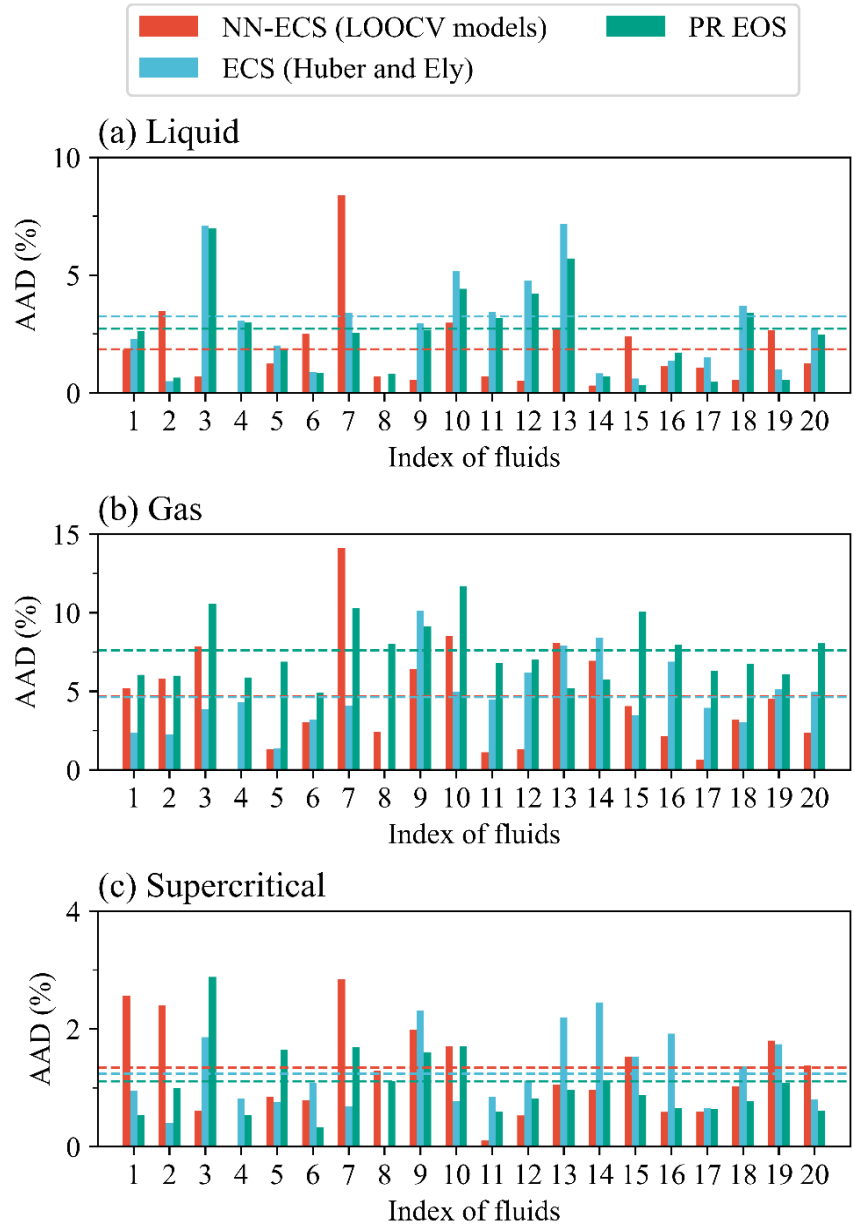

Fig. 5. AAD of the residual enthalpy for each fluid across different phases: (a) Liquid, (b) gas, (c) Supercritical. The dashed lines indicate the AAD in total dataset.

The AAD of the proposed NN-ECS for HFO and HFC across different phases is illustrated in Fig. 6. The similar AAD values between HFO and HFC indicate that the model exhibits no inherent bias toward either compound, demonstrating its balanced predictive capability for both HFO and HFC. For density, the gas phase exhibits the lowest AAD, while the supercritical phase shows the highest. For residual entropy and residual enthalpy, the AAD values of liquid and supercritical phases are comparable, whereas the gas phase yields the largest deviations. Reasons for the significant errors in gaseous residual energy properties will be analyzed in section 4.1, and they are demonstrated to have negligible impact on the total energy properties calculation.

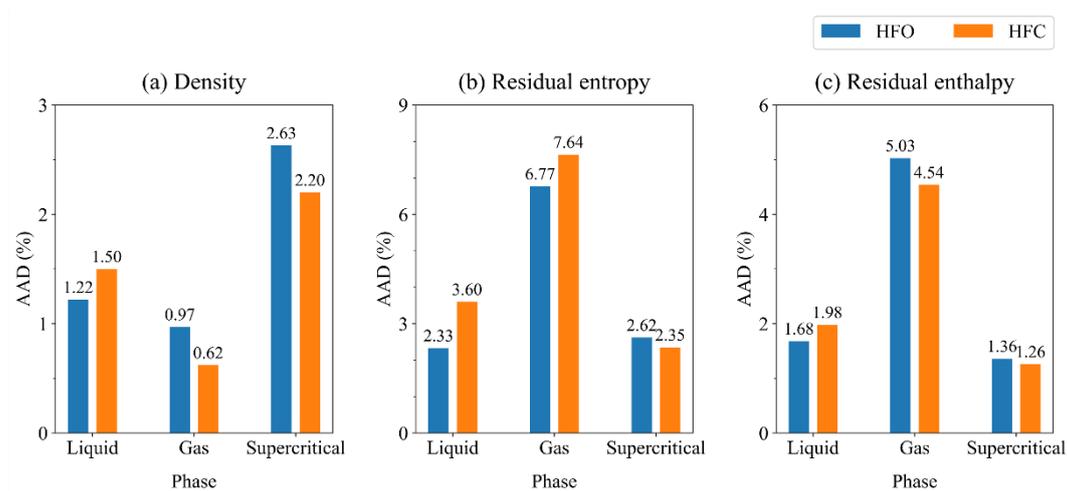

Fig. 6. AAD of the proposed NN-ECS for HFO and HFC across different properties: (a) Density, (b) Residual entropy, (c) Residual enthalpy.

*3.3 Application to three new HFO refrigerants*

Three new HFO refrigerants (R1132(E), R1336mzz(E), and R1354mzy(E)) were adopted to further validate the reliability of the NN-ECS model. Their experimental density data were compared with the predicted values, and the results are given in Table

10. For R1132(E), its critical parameters were adopted from the publication of its newly developed Helmholtz EOS [13], with $T_c$ = 348.82 K and $\rho_c$ = 6.793 mol·L$^{-1}$. Its *PVT* data of Sakoda et al. [50] are calculated with an AAD of 1.91%. For R1336mzz(E), its critical parameters were also adopted from the publication of its newly developed EOS [11], with $T_c$ = 403.53 K and $\rho_c$ = 3.129 mol·L$^{-1}$. Two sets of *PVT* data are available for R1336mzz(E), but minor inconsistencies exist between them. The data from Tanaka et al. [51] show better agreement with the NN-ECS model, with an AAD of 1.29%, and the data of Sakoda et al. [52] show AAD of 2.88%. For R1354mzy(E), its $T_c$ was adopted from the measurements of Kimura et al. [53] with the value of 424.73 K. Regarding $\rho_c$, we employed a value of 3.436 mol·L$^{-1}$ estimated by a quantitative structure-property relationship (QSPR) model fitted from the refrigerants in REFPROP. Its *PVT* data of Kimura et al. [54] are calculated with AAD of 0.87%. The good agreement between the predicted and experimental density values further demonstrates the generalization capability of the proposed NN-ECS model.

Table 10 AAD (%) between the experimental and NN-ECS predicted density values for three new HFO refrigerants.

|  | N* | Liquid | Vapor | Supercritical | All |
| --- | --- | --- | --- | --- | --- |
| R1132(E) | | | | | |
| Sakoda et al. [50] | 58 (51) | 1.79 | 1.78 | 2.48 | 1.91 |
| R1336mzz(E) | | | | | |
| Tanaka et al. [51] | 156 (154) | 1.71 | 0.95 | 1.28 | 1.29 |

| Sakoda et al. [52] | 39 (39) | 3.86 | 1.37 | 3.08 | 2.88 |

R1354mzy(E)

| Kimura et al. [54] | 102 (102) | 1.08 | 0.68 | N/A | 0.87 |

*The values in parentheses indicate the number of data points used for comparison after excluding the near-critical region.

## 4. Discussion

*4.1 Reasons for the large deviations in gas residual energy properties*

The NN-ECS model, PR EOS, and the conventional ECS model using R134a as reference fluid, all show relatively large deviations for the residual entropy and enthalpy in the gas phase. This phenomenon can be explained by the thermodynamic characteristics of the gas phase: ideal gas contribution dominates the energy properties, while the residual contribution accounts for only a minimal portion due to the extremely weak intermolecular interactions. Taking R1234yf as an example, Fig. 7(a) and (b) display the values of $|s^R/s|$ and $|h^R/h|$ in the pressure-temperature diagram, respectively. For residual entropy, the average value of $|s^R/s|$ is 1.11% in the entire gas phase. $|s^R/s|$ reaches its minimum in the low-temperature and low-pressure gas phase, which is the primary source of large prediction deviations. The value of $|s^R/s|$ increases with the vapor density, peaking at 6.53% as very close to the critical point, where the vapor density also reaches its maximum. As $|s^R/s|$ increases, the $s^R$ prediction deviation decreases correspondingly in this region. The trend of $|h^R/h|$ is the same as that of $|s^R/s|$, but the difference is that the proportion of residual enthalpy is slightly larger than that of residual entropy, with a mean and maximum values of 3.69% and 20.82%, respectively. Similarly, the $h^R$ prediction deviations also decrease with increasing proportion. Given the value of total energy properties to be correct, minor deviations in the ideal gas part would lead to significant variation in the residual part. Thus, the reference value of the energy properties in gas phase is less accurate, and this is the reason why the models show relatively large deviations for residual energy properties in the gas phase.

In the gas region, the residual energy contribution only accounts for merely a few percent or less. Although the residual proportion increases as the increasing of gas density near the critical point, the prediction deviation decreases correspondingly. Therefore, it is concluded that the large deviations in the residual part have a negligible effect on the total energy calculations. For an accurate calculation of energy properties in the gas phase, the ideal gas contribution, rather than the residual contribution, must be precisely known. The ideal gas contribution, is usually derived from $c_p^{id}$, which could be determined by the statistic mechanics method [55] or sound speed measurements in the gas phase [9].

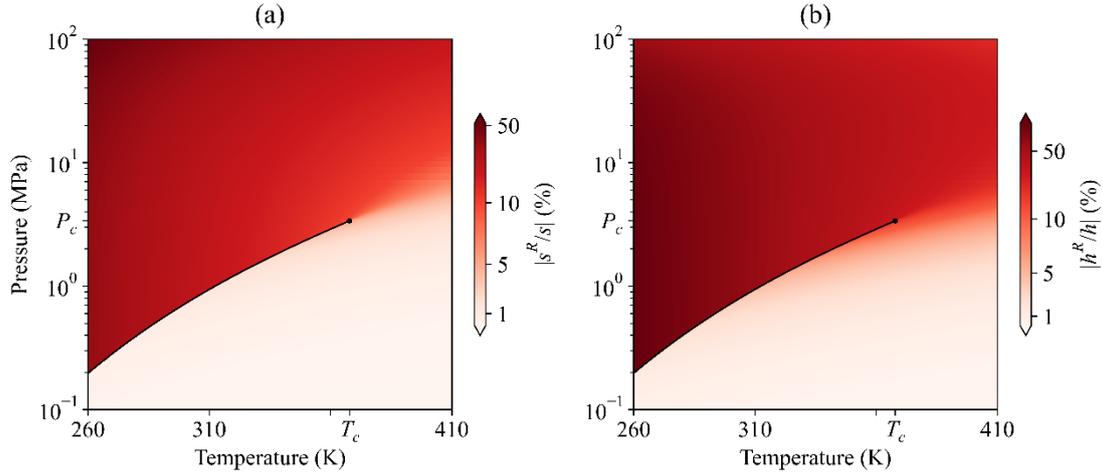

Fig. 7. Plot of the proportion of the residual energy properties for R1234yf: (a) $|s^R/s|$, (b) $|h^R/h|$.

*4.2 Influence of the uncertainty in critical parameters*

The proposed NN-ECS predicted the residual thermodynamic properties of fluorine-containing refrigerants from their molecular structures; in addition, their $T_c$ and $\rho_c$ should also be known in advance to turn the temperature and density into reduced forms. The critical parameters could be estimated using QSPR methods [56; 28] or directly determined by experimental measurement. However, uncertainty, particularly for $\rho_c$, is introduced into the model whether using the estimation or experimental methods. Taking R1234yf as an example, Fig. 8 shows the variations in calculated values of density, residual entropy, and residual enthalpy by introducing a 1% uncertainty in $T_c$ or $\rho_c$. For the density, the uncertainty is stable in the region far from the critical point, with values typically less than 1%. The uncertainty increases dramatically when approaching the critical point, especially in the supercritical region.

This is caused by the singularity of critical point, where the isothermal compressibility and isobaric heat capacity diverge to infinity and the sound speed tends to zero. The huge compressibility in this region will result in drastic density changes with small temperature or pressure changes, making the ECS method not applicable to describe the density behavior in this region. Regarding the residual entropy and residual enthalpy, the effect of the critical point is not obvious although the $c_p$ diverges. The reason is that the residual enthalpy is obtained through the integration of $c_p$ and the divergency only occurs in a narrow region, thus the residual entropy and enthalpy remain flat in this region. In the liquid and supercritical regions, the uncertainty in $\rho_c$ affects the calculated values of residual entropy more significantly than that of $T_c$. The situation turns opposite for residual enthalpy, where $T_c$ has a much larger effect than $\rho_c$. This conclusion is consistent with the results in Table 8 and Table 9: for the NN-ECS (whether the full-data model or LOOCV models), the AAD of the residual entropy is larger than that of residual enthalpy, owing to the fact that uncertainty in reference value of $\rho_c$ is much larger than that in $T_c$ in REFPROP.

Furthermore, the uncertainty in the calculated values of density, residual entropy, and residual enthalpy, with an uncertainty of 1% in $T_c$ or $\rho_c$, are calculated and averaged for all fluids in the total dataset, excluding the region near the critical point. The results in the total dataset are presented in Table 11, and showing trends largely consistent with those for R1234yf in Fig. 8.

Table 11 Uncertainty (%) in density, residual entropy, and residual enthalpy with an uncertainty of 1% in $T_c$ or $\rho_c$ in the total dataset.

|  | Density | Residual entropy | Residual enthalpy |
|---|---|---|---|
| Liquid | | | |
| $T_c$ | 0.71 | 0.39 | 2.17 |
| $\rho_c$ | 0.94 | 1.62 | 0.73 |
| Gas | | | |
| $T_c$ | 0.68 | 2.13 | 2.37 |
| $\rho_c$ | 0.23 | 0.97 | 0.96 |
| Supercritical | | | |
| $T_c$ | 3.14 | 1.01 | 2.10 |
| $\rho_c$ | 0.70 | 1.05 | 0.55 |

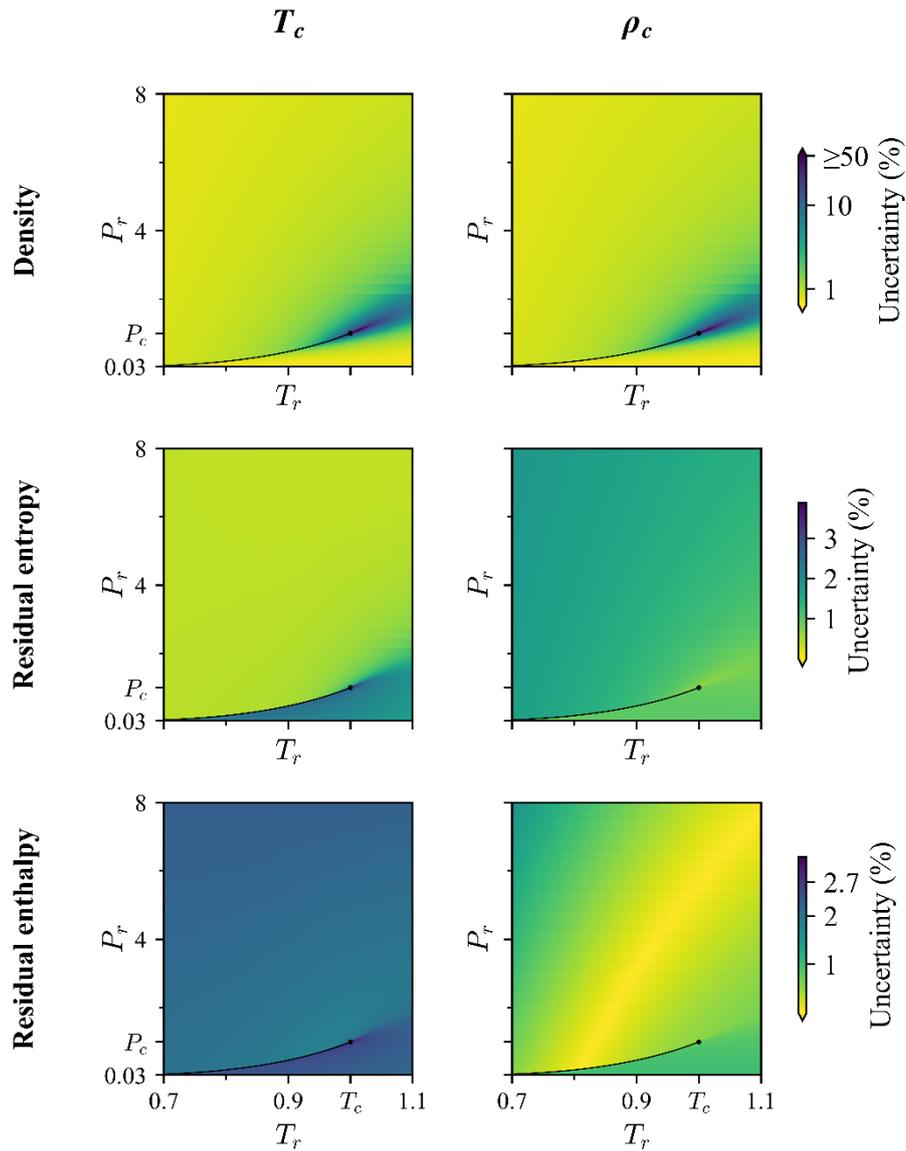

Fig. 8. Variation in the calculated thermodynamic properties of R1234yf under 1% uncertainty in critical parameters. Rows correspond to density, residual entropy, and residual enthalpy; columns correspond to $T_c$ and $\rho_c$. Darker colors indicate larger variations. The saturation line and critical point are marked with black curve and dot, respectively.

*4.3 Strengths, limitations, and application ranges of the proposed NN-ECS*

Compared with the conventional ECS methods or the PR-EOS, the proposed NN-ECS model achieves significantly improved accuracy for the density and energy properties in liquid and supercritical phases. The possible reasons are as follows:

- The shape factors in the ECS method are modeled using a neural network, thereby avoiding the potential limitations introduced by the empirical formulas. In addition, leveraging the powerful fitting capability of neural networks and the substantial amount of available data, 44 fluids were used in the NN-ECS model training, whereas other models only employed fewer than 10 fluids.

- Utilizing the flexibility of neural networks, NN-ECS model incorporates the information of microscopic molecular structure similarities by embedding a GNN module, whereas conventional models rely solely on macroscopic critical parameters and acentric factor to characterize the fluids.

- As shown Eq. (5), in the original ECS theory, $\rho_c$ should be incorporated in the scaling factor $h_j$. Previous ECS models circumvented the use of $\rho_c$ by introducing the critical compressibility factor in scaling factor $h_j$. Consequently, the previous ECS models employ $T_c$ and $P_c$, while the NN-ECS model employs $T_c$ and $\rho_c$, which makes the NN-ECS more consistent with the original ECS theory.

The ECS method with formulations proposed by Huber and Ely, is further demonstrated to be accurate when directly fitted to the vapor pressure and saturated liquid density of studied fluid, as evidenced by the results of ECS model with fluid-specific parameters in Group 1. However, this method shows relatively large deviations when using universal parameters. For fair comparisons, we have redetermined the universal parameters based on our dataset with R1234ze(E) as the reference fluid. These recalculated universal parameters (adjusted to the 44 fluids in this work) are close to those reported by Teraishi et al. (adjusted to HFO and HCFO), as shown in Table 5. Furthermore, their predictive performance is also similar, as evidenced by the results in Tables 7-9. This result may be due to the fact that only four universal parameters are insufficient to describe a large number of fluids.

Although the proposed NN-ECS achieves high accuracy for single phase properties, its limitation is the poor vapor pressure prediction capability. Regarding vapor pressure, the PR EOS or the ECS model proposed by Teraishi et al. provides more accurate predictions. The reasons are as follows:

- The PR EOS and ECS model of Teraishi et al. are all fitted from the vapor pressure data; moreover, critical pressure ($P_c$) and acentric factor of fluids are used in their models, this means that two points in the saturation curve are already known in advance. This is the reason why PR EOS and ECS model of Teraishi et al. can accurately predict the vapor pressure of fluids.

- The critical temperature and critical density of fluids were used in the NN-ECS; thus, the model has no information about the saturation curve in advance. In addition, different from the explicit ECS model of Teraishi et al., the NN-ECS model in this work obtains vapor pressure using the criterion in Eq. (13), thus the calculated values are more sensitive to the precision of Helmholtz energy surface, and a slight inconsistency in the surface may have a great impact on the calculation result. However, NN-ECS have AAD of less than 5% for most fluids except a few fluids, and this result is acceptable considering that it does not know the $P_c$ and $\omega$ in advance.

Considering the molecules in the dataset, the data range used for model training, and the strengths, limitations of the proposed model, the NN-ECS is applicable to the following compounds:

(1) Any straight-chain HFO containing only one double bond.

(2) Any straight-chain HFC with more than two carbon atoms.

Since all HFO and HFC in the dataset fall into the above two categories, the performance of the NN-ECS cannot be guaranteed for compounds beyond this range. Nevertheless, this range still covers a sufficiently large chemical space that is enough for the purpose of searching for new HFO refrigerants.

The applicable range spans from $0.7T_c$ to $1.1T_c$ in temperature and from 0.1 MPa to 50 MPa in pressure, which has covered the common operating conditions for refrigerants. Within this range, the proposed NN-ECS can accurately predict the single-phase density (excluding the near-critical region), residual entropy, and residual enthalpy of compounds in the above-mentioned two categories, whereas the PR EOS or the ECS model of Teraishi et al is preferred for the vapor pressure calculations.

## 5. Conclusion

In this work, we introduced a neural-network extended corresponding state (NN-ECS) model to predict the residual thermodynamic properties of fluorine-containing refrigerants, especially for HFO, which is the most promising new environmentally friendly refrigerants. The model was trained using the highly accurate data of existing fluids in the REFPROP database, and its predictive capability is rigorously evaluated using the LOOCV method.

Compared with conventional ECS models or the cubic EOS, the strength of NN-ECS is its significantly improved accuracy for the properties in the single phase, particularly for the density and residual entropy in liquid and supercritical phases. However, the current limitation of NN-ECS is its poor vapor pressure prediction capability, which essentially stems from the way it was constructed. For the vapor pressure calculations, the PR EOS or the ECS model of Teraishi et al. is recommended for their high accuracy.

The influence of the uncertainty in $T_c$ and $\rho_c$ on the calculated thermodynamic properties was also analyzed. Results show that the proximity to the critical point significantly affects the calculated density, $\rho_c$ is the dominant factor on residual entropy, while $T_c$ is the dominant factor on residual enthalpy. In future work, we plan to develop reliable prediction models for the ideal gas Helmholtz energy and critical parameters (particularly $\rho_c$) of HFO refrigerants to achieve the complete prediction of their thermodynamic properties. The proposed NN-ECS is expected to play a pivotal role in the assessment and discovery of potential new refrigerants.

**Acknowledgements**

This work is supported by the National Natural Science Foundation of China (Grant No.: 52176171).

**Data availability**

The model code (full-data version) along with a usage example is publicly available on GitHub repository (https://github.com/WangGangustc/NN_ECS.git).